\documentclass{article}
\usepackage[dblblindworkshop, final]{neurips_2025}
\usepackage{natbib}
\usepackage[utf8]{inputenc} 
\usepackage[T1]{fontenc}    
\usepackage[dvipsnames]{xcolor}
\usepackage[colorlinks=true, citecolor=RoyalBlue, linkcolor=RoyalBlue, urlcolor=Black]{hyperref}
\usepackage{url}           
\usepackage{booktabs}      
\usepackage{amsfonts}       
\usepackage{nicefrac}       
\usepackage{microtype}      
\microtypesetup{tracking=true}
\usepackage{placeins}
\usepackage{amsmath}
\usepackage{dsfont}
\usepackage{bm}  

\usepackage{graphicx}
\usepackage{mathrsfs}
\usepackage{enumitem}
\usepackage{multirow} 

\usepackage{array}          
\usepackage{makecell}       
\usepackage[table]{xcolor}  
\usepackage{geometry}      
\usepackage{caption} 
\usepackage{colortbl}
\usepackage{dblfloatfix}
\usepackage{amssymb}
\usepackage{pifont}
\usepackage{fontawesome5}
\usepackage{tcolorbox}
\usepackage{nicematrix}
\usepackage[mathscr]{euscript}
\usepackage{soul}
\usepackage{fancyvrb}
\definecolor{mark_gray}{gray}{0.9}
\sethlcolor{mark_gray}
\newcommand{\gcell}[1]{\cellcolor{lightgraybgDarker}{\raisebox{-0.3ex}[0pt][0pt]{#1}}}
\usepackage{arydshln} 
\definecolor{darkred}{rgb}{0.6, 0, 0}
\definecolor{darkgreen}{rgb}{0, 0.3, 0}
\definecolor{lightgraybg}{gray}{0.95}
\definecolor{lightgraybgDarker}{gray}{0.90}
\definecolor{lightgrayedge}{gray}{0.75}

\newcommand{\bad}[1]{
  \begingroup
  \setlength{\fboxsep}{1pt}
  \fcolorbox{orange}{white}{\textcolor{black}{#1}}%
  \endgroup
}
\definecolor{marker_gray}{gray}{0.5}
\newcommand{\greenOne}[1]{\begingroup\setlength{\fboxsep}{0.6pt}\colorbox{green!20}{#1}\endgroup}
\newcommand{\greenTwo}[1]{\begingroup\setlength{\fboxsep}{0.6pt}\colorbox{cyan!20}{#1}\endgroup}
\newcommand{\greenThree}[1]{\begingroup\setlength{\fboxsep}{0.6pt}\colorbox{yellow!20}{#1}\endgroup}
\newcommand{\greenFour}[1]{\begingroup\setlength{\fboxsep}{0.6pt}\colorbox{orange!20}{#1}\endgroup}
\newcommand{\greenFive}[1]{\begingroup\setlength{\fboxsep}{0.6pt}\colorbox{purple!20}{#1}\endgroup}

\usepackage{CJKutf8}
\newcommand{\nen}{\begin{CJK*}{UTF8}{gbsn}年\end{CJK*}}
\setlist[itemize]{left=0cm}

\title{Multimodal Survival Analysis with Locally Deployable Large Language Models}

\author{%
  Moritz Gögl\\
  University of Oxford\\
  \texttt{moritz.gogl@keble.ox.ac.uk} \\
  \And
  Christopher Yau \\
  University of Oxford \\
  \texttt{christopher.yau@wrh.ox.ac.uk} \\
}

\begin{document}
\maketitle
\begin{abstract}
We study multimodal survival analysis integrating clinical text, tabular covariates, and genomic profiles using locally deployable large language models (LLMs). As many institutions face tight computational and privacy constraints, this setting motivates the use of lightweight, on‑premises models. Our approach jointly estimates calibrated survival probabilities and generates concise, evidence‑grounded prognosis text via teacher–student distillation and principled multimodal fusion. On a TCGA cohort, it outperforms standard baselines, avoids reliance on cloud services and associated privacy concerns, and reduces the risk of hallucinated or miscalibrated estimates that can be observed in base LLMs.
\end{abstract}

\section{Introduction}
Survival analysis estimates the probability of an event over time and is central to medical decision-making (e.g., forecasting mortality or disease progression). Classical models operate on structured covariates (e.g., age, sex, genomic data), while clinical practice also generates rich unstructured data, such as clinical reports. Recent large language models (LLMs) can reason over such text and produce human-readable assessments, but cloud-hosted models raise privacy concerns and heavyweight local deployments are impractical for many institutions \cite{kingsfundInfrastructureInnovation}.
Moreover, base LLMs are not calibrated for survival prediction as they are not trained on raw survival data; they typically recall published summary statistics–and may hallucinate–rather than producing data‑grounded estimates \cite{xiong2024}.

We present a unified, locally deployable multimodal survival framework that pairs a compact causal LLM with structured covariates and gene expression. Our model jointly produces calibrated survival curves and concise prognosis explanations. A teacher–student pipeline first queries a large teacher LLM for numeric survival probabilities at fixed horizons and a brief assessment; the student then learns from both the teacher’s verbalized reasoning and the observed survival outcomes. The architecture supports either a discrete-time hazards model or a Cox proportional hazards (CoxPH) \cite{Cox1972} model, and fuses modalities by concatenation or via separate gated heads. Compared to prior multimodal survival models \cite{Zhao2021_BERTSurv, Song2025}, our approach couples survival estimation with concise explanations while remaining lightweight and locally deployable. 
\vspace{-0.16cm}
\paragraph{Contributions.}
Our work makes three contributions: (1) a calibrated locally deployable multimodal survival framework that couples a compact causal LLM (1.5B parameters) with covariates and gene expression, supporting both discrete and CoxPH heads with flexible fusion; (2) a teacher–student fine-tuning scheme with a single forward-pass text objective that distills numeric survival probabilities and rationales; and (3) an empirical evaluation on a TCGA cohort \cite{TCGA2013} showing improved performance over baselines, alongside concise, verbalized prognosis explanations.

\section{Related Work}
\vspace{-0.1cm}
\label{sec:related_work}
Methods in survival analysis range from parametric and semiparametric models to nonparametric estimators, with modern ensemble and deep‑learning approaches that learn flexible risk and event‑time structures. Cox models \cite{Cox1972} remain standard but assume a log‑linear covariate effect and proportional hazards, while Random Survival Forests \cite{Ishwaran2008_RSF} relax such assumptions. Deep learning introduced end‑to‑end models that infer complex survival patterns: DeepSurv \cite{Katzman2018_DeepSurv} replaces the Cox linear predictor with a neural network, DeepHit \cite{Lee2018_DeepHit} models discrete hazards and competing \nolinebreak risks, \nolinebreak and \nolinebreak Neural Survival Clustering \cite{Jeanselme2022_NSC} learns mixture structure in survival data. Leveraging text for survival prediction emerged with BERTSurv \cite{Zhao2021_BERTSurv}, which uses transformer embeddings of clinical notes to improve concordance over tabular baselines, employing a BERT‑based backbone \cite{Devlin2018_BERT} (e.g., ClinicalBERT \cite{Huang2019_ClinicalBERT}). A recent survey by Jeanselme et al.~\cite{jeanselme2024} reviews language‑model strategies for survival, covering direct prompting, feature extraction, and fine‑tuning pipelines, and highlights open issues around censoring and evaluation protocols. Foundation models have also been explored for multimodal survival: Song et al.~\cite{Song2025} show that zero‑shot embeddings from foundation models can be combined with classical survival models to yield gains over unimodal baselines, and discuss risks of hallucination in text summarization. Complementary work applies LLMs directly to pathology reports for cancer type, stage, and prognosis assessment \cite{Saluja2025}, focusing on text‑only predictions without calibrated survival curves. Moreover, our framework draws motivation from the approach introduced in \cite{Strong_2025}, which combines calibrated hidden‑state and verbalized signals in the context of guided deferral systems.

\section{Methods}
\vspace{-0.1cm}
\subsection{Problem Setup and Overview}
\begin{figure}[h!] 
    \centering
    \includegraphics[width=0.9\textwidth]{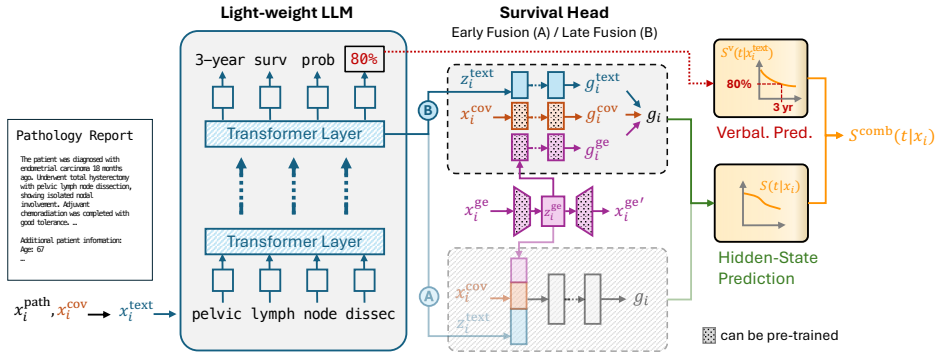}
    \caption{Overview of the proposed framework. A compact causal LLM encodes the input text into hidden embeddings used by a survival model (here, CoxPH) and produces verbalized survival estimates with explanatory text. Covariates and gene expression are fused either early (A) or late (B).}
    \label{fig:architecture}
\end{figure}

We consider right-censored survival data consisting of triples $x_i=(x^{\text{path}}_i, x^{\text{cov}}_i, x^{\text{ge}}_i)$ and outcomes $(t_i, e_i)$ for samples $i=1, \dots, N$, where $t_i$ denotes the observed follow-up time and $e_i\in\{0,1\}$ is the event indicator ($e_i=1$ indicating death, $e_i=0$ indicating censoring). Here, $x^{\text{path}}_i$ is a free-text pathology report; $x^{\text{cov}}_i\in\mathbb{R}^{d_\text{c}}$ are tabular covariates (e.g., age, sex, cancer type); and $x^{\text{ge}}_i\in\mathbb{R}^{d_\text{g}}$ is a high-dimensional gene-expression vector. For the text channel, we form a combined input $x^{\text{text}}_i$ by appending a formatted patient-info string to the report. We estimate the conditional survival distribution $S(t\mid x)$ in two complementary ways (Fig.~\ref{fig:architecture}). First, a hidden-state pathway encodes the text with a compact causal LLM whose representation feeds a survival head (either discrete-time hazards or CoxPH). Tabular covariates and gene-expression latents are fused via early concatenation or late fusion with gated heads. Second, a verbalized pathway has the LLM generate an explicit survival probability together with a concise rationale, which we map to a full survival curve.

\subsection{Hidden-State Survival Predictions}
We use hidden representations from a compact causal LLM together with covariates and gene expression to produce well-calibrated survival estimates:
\paragraph{Text encoder and pooling.} A causal LLM encodes the tokenized pathology report into hidden states $H \in \mathbb{R}^{L \times d}$. We form a fixed-size embedding $z^{\text{text}}$ via simple self-attention pooling, which aggregates token representations by similarity and averages them into a sequence-level vector:
\begin{equation}
A = \operatorname{softmax}(H H^{\top}),\quad \tilde{H} = A H,\quad z^{\text{text}} = \frac{1}{L} \sum_{\ell=1}^L \tilde{H}_{\ell:}.
\end{equation}

\paragraph{Gene-expression autoencoder.} High-dimensional gene expression (GE) is compressed with an autoencoder $\text{Dec}(\text{Enc}(x^{\text{ge}}_i))$, yielding a latent $z^{\text{ge}}_i=\text{Enc}(x^{\text{ge}}_i)$. The reconstruction objective 
$\mathcal{L}_{\text{AE}}\!=\!\!\!\,\mathbb{E}_i[\lVert \text{Dec}(\text{Enc}(x^{\text{ge}}_i)\!)\!-\!x^{\text{ge}}_i \rVert_2^2]/d_{\text{g}}$ 
regularizes the latent while preserving survival-relevant signal \cite{Tong2020}.
\vspace{-0.05cm}
\paragraph{Survival models.} We support two survival network heads: (i) a \emph{discrete-time hazards} model \cite{Gensheimer2019, Kvamme2021_ContAndDisc}, which outputs logits $o\in\mathbb{R}^{B}$ over $B$ time bins (hazards $h_b=\sigma(o_b)$, survival $S(t_b)=\prod_{k\le b}(1-h_k)$) and is trained with a masked Bernoulli objective; and (ii) \emph{CoxPH} \cite{Katzman2018_DeepSurv}, which outputs a log-risk score $g$ and is optimized via the negative partial log-likelihood. The corresponding training objectives are
\begin{equation}
\mathcal{L}_{\text{disc}} = \frac{\sum_{i,b} a_{ib}\, \mathrm{BCE}(h_{ib}, y_{ib})}{\sum_{i,b} a_{ib}} \qquad \mathcal{L}_{\text{cox}} = -\frac{1}{\sum_i e_i}\sum_{i: e_i=1} \Big( g_i - \log \sum_{j: t_j \ge t_i} e^{g_j} \Big).
\end{equation}
Here, $y_{ib}=\mathds{1}\{e_i=1,\; t_{b-1}<t_i\le t_b\}$ indicates an event for individual $i$ in bin $b$, and $a_{ib}=\mathds{1}\{\text{at risk at the start of $b$}\}$ masks at-risk samples that have not been censored or died by time bin $b$.

\paragraph{Fusion strategies.} We consider two regimes to integrate text, covariates, and gene expression \cite{Gadzicki2020}:
\begin{enumerate}[label=(\Alph*),leftmargin=*,itemsep=0pt,topsep=0pt]
\item Early fusion concatenates $z=[z^{\text{text}}; x^{\text{cov}}; z^{\text{ge}}]$ and feeds a single head $f$, enabling rich cross-modal interactions at the cost of tighter coupling.
\item Late fusion learns modality-specific heads $f^{\text{text}}, f^{\text{cov}}, f^{\text{ge}}$ and combines their outputs with learned gates $\gamma^{\text{text}}$, $\gamma^{\text{cov}}$, $\gamma^{\text{ge}}$ (see Appendix~\ref{app:latefusion}). This enables separate \emph{pre-training} of $f^{\text{cov}}$ and $f^{\text{ge}}$ \pagebreak[2] without being constrained by the larger memory footprint of end-to-end LLM fine-tuning.
\end{enumerate}

\subsection{Verbalized Survival Prediction and Assessment}
\begin{figure}[t!] 
    \centering
    \includegraphics[width=1\textwidth]{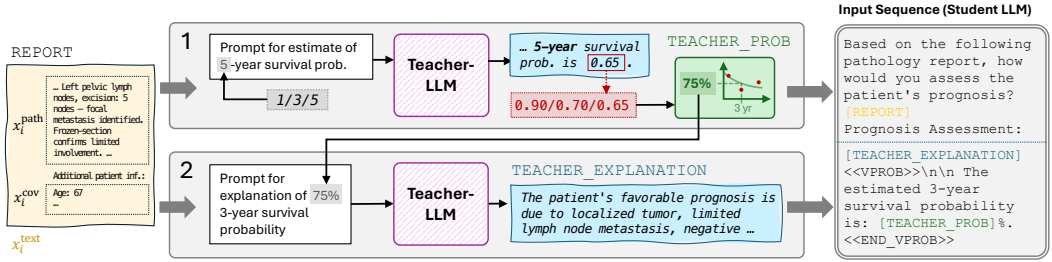}
    \caption{Teacher pipeline for constructing the student target. (1) Numeric prompting at 1/3/5 years from the input text; extract survival probabilities, and fit a parametric curve (exponential by default) to obtain the 3‑year value. (2) Explanation prompting conditioned on the text and the rounded 3‑year percentage; the explanation plus a marked probability sentence form the student’s training target.}
    \label{fig:teacher_pipeline}
    \vspace{-0.05cm}
\end{figure}

Inspired by an approach introduced for instruction-tuned LLMs in guided deferral systems \cite{Strong_2025}, we additionally use the generative capabilities of the same compact causal LLM to produce a concise prognosis explanation and an explicit 3-year survival probability statement.
\paragraph{Teacher–student distillation.} As illustrated in Fig.~\ref{fig:teacher_pipeline}, we first query a larger \emph{teacher} LLM (here: DeepSeek-R1 Distill Qwen-32B \cite{DeepSeek2025}) offline with two prompts: (1) a sequence of numeric-only instructions to return survival probabilities at 1/3/5 years conditioned on $x^\text{text}$; and (2) an explanation prompt conditioned on $x^\text{text}$ and the rounded 3-year survival probability, \texttt{TEACHER\_PROB}, computed from an exponential fit of the extracted numeric predictions (see Appendix~\ref{app:teacher_pipeline}). The generated teacher explanation and probability estimate are then used to construct the target sequence: 
\hl{\texttt{[TEACHER\_EXPLANATION] $\text{<<}$VPROB$\text{>>}$$\backslash$n$\backslash$n The estimated 3-year survival probability is: [TEACHER\_PROB]\%.$\text{<<}$END\_VPROB$\text{>>}$}},
which is learned by the \emph{student} model (here: DeepSeek-R1 Distill Qwen-1.5B \cite{DeepSeek2025}) during training, conditioned on $x^\text{text}$.

For our use case, the verbalized prediction sentence–delimited by \texttt{<<VPROB>>} and \texttt{<<END\_VPROB>>}–and, within it, the numeric probability, \texttt{TEACHER\_PROB}, are the most critical parts of the assessment. We therefore upweight the cross-entropy on tokens in this span and on the numeric substring using weights $w$ and $w^{\text{num}}$, yielding the following loss:
\begin{equation}
\mathcal{L}_{\text{text}} \;=\; \mathcal{L}^{\text{full}}_{\text{text}} \; + \; (w{-}1)\,\mathcal{L}^{\text{vprob}}_{\text{text}} \; + \; (w^{\text{num}}{-}1)\,\mathcal{L}^{\text{num}}_{\text{text}}.
\end{equation}
\paragraph{Calibration correction} By default, all samples contribute to the text loss. However, since teacher predictions are obtained by prompting an LLM rather than learned directly from observed survival outcomes, they may be miscalibrated. Following a similar idea to \cite{Strong_2025}, we optionally mask out text loss contributions for samples whose teacher 3-year survival estimate, \texttt{TEACHER\_PROB}, is inconsistent with the observed status at the assessment horizon: (i) the event occurred before 3 years yet the teacher assigns a high survival probability (>50\%), or (ii) the \nopagebreak[4] individual is known to be alive/at-risk at 3 years yet the teacher assigns a low survival probability (<50\%). All samples remain in the dataset and fully contribute to survival objectives; only language-model targets are masked to reduce the influence of potentially miscalibrated supervision in text generation.

\subsection{Objective and Optimization}
\vspace{-0.05cm}
We jointly optimize the survival model, gene-expression reconstruction, and text generation, reweighting the latter two with coefficients $\alpha$ and $\beta$. The total loss is then given by
\begin{equation}
\mathcal{L} = \mathcal{L}_{\text{surv}} + \alpha\,\mathcal{L}_{\text{AE}} + \beta\,\mathcal{L}_{\text{text}},
\end{equation}
where $\mathcal{L}_{\text{surv}}\in\{\mathcal{L}_{\text{disc}},\mathcal{L}_{\text{cox}}\}\!$ depending on the selected survival model. 
At test time, we compute the survival function $S(t|x_i)$ (probability of surviving beyond $t$) from the survival-head outputs, and fit an exponential curve $S^{\text{v}}(t|x_i^{\text{text}})$ from the verbalized 3-year survival probability. We then form a convex combination $S^{\text{comb}}(t|x_i)=(1-\lambda)\,S(t|x_i)+\lambda\,S^{\text{v}}(t|x_i^{\text{text}})$, with $\lambda\in\![0,1]$ selected on the validation set to maximize concordance \cite{Strong_2025}.\footnote{If no probability can be extracted, we set the combined prediction to the hidden-state curve $S^{\text{comb}}(t|x_i)=S(t|x_i)$, and mean-impute $S^{\text{v}}(t|x_i^{\text{text}})=\overline{S^{\text{v}}(t|x^{\text{text}})}$ across test samples for verbalized-only evaluation.}

\section{Experiments}
\vspace{-0.1cm}
\subsection{Experimental Setup}
\vspace{-0.05cm}
We evaluated on a TCGA-derived cohort \cite{TCGA2013} of 8,902 samples comprising pathology reports, tabular covariates, and gene expression; details are provided in Appendix~\ref{app:data}. Across both survival models (discrete and CoxPH), we ablate fusion strategy (early vs.~late), optional pre-training of covariate- and gene-expression-specific survival heads under increased batch size, calibration correction (CC), and a variant conditioned only on $x^{\text{text}}$. Baselines include BERTSurv \cite{Zhao2021_BERTSurv} and unimodal experiments on covariates and gene expression under both survival models. We additionally report our teacher’s verbalized prediction performance, deriving survival functions via exponential fitting. We report time-dependent concordance ($C^{\text{td}}$) and integrated Brier score (IBS); definitions are given in Appendix~\ref{app:metrics}.

\subsection{Results}
\vspace{-0.05cm}
Table~\ref{table:results} summarizes performance across configurations; our standard setting (late fusion, no pre-training, no calibration correction) is highlighted in gray. Across settings, hidden-state and combined predictions outperformed baselines, confirming gains from multimodal fusion; hidden-state predictions were consistently stronger than verbalized ones, and blending the two yielded modest, consistent improvements in $C^{\text{td}}$ and IBS. The teacher achieved substantially better verbalized performance, reflecting its larger model capacity. Moreover, its prediction was obtained via three separate prompts, making it more robust. Late fusion generally improved hidden-state and combined performance; a variant conditioned only on $x^{\text{text}}$ underperformed, consistent with the strong predictive signal in gene expression observed in baselines. Calibration correction left CoxPH unchanged but markedly improved the discrete model’s performance, yielding the best overall discrimination. Pre-training modality-specific heads with larger batches to overcome memory constraints under end-to-end LLM training yielded no improvement. Among baselines, CoxPH on gene expression performed best and exceeded BERTSurv, even though BERTSurv leverages information from all modalities.
\begin{table}[t!]
\centering
\caption{Performance across configurations and baselines.}
\setlength{\tabcolsep}{3pt}
\begin{footnotesize}
\begin{tabular}{lccccccc|cc|cc|cc}
\toprule
\multicolumn{8}{c|}{Model/Configuration} & \multicolumn{2}{c|}{Hidden-state} & \multicolumn{2}{c|}{Verbalized} & \multicolumn{2}{c}{Combined} \\
\cmidrule(l){1-8} \cmidrule(l){9-10} \cmidrule(l){11-12} \cmidrule(l){13-14}
Name & Text$^{*}$ & Cov & GE & Survival & Fusion & Pretrain & CC & $C^{\text{td}}\!\uparrow$ & IBS $\!\downarrow$ & $C^{\text{td}}\!\uparrow$ & IBS $\!\downarrow$ & $C^{\text{td}}\!\uparrow$ & IBS $\!\downarrow$ \\
\midrule
BERTSurv & \faCheck & \faCheck & \faCheck & CoxPH & Early & \textemdash{} & --- &  0.691 &  0.150 & --- & --- & --- & --- \\
Cov-only & \textcolor{lightgray}{\faTimes} & \faCheck & \textcolor{lightgray}{\faTimes} & Discrete & --- & \textemdash{} & --- &  0.665 &  0.149 & --- & --- & --- & --- \\
& \textcolor{lightgray}{\faTimes} & \faCheck & \textcolor{lightgray}{\faTimes} & CoxPH & --- & \textemdash{} & --- &  0.668 &  0.149 & --- & --- & --- & --- \\
GE-only & \textcolor{lightgray}{\faTimes} & \textcolor{lightgray}{\faTimes} & \faCheck & Discrete & --- & \textemdash{} & --- &  0.734 &  0.139 & --- & --- & --- & --- \\
& \textcolor{lightgray}{\faTimes} & \textcolor{lightgray}{\faTimes} & \faCheck & CoxPH & --- & \textemdash{} & --- &  0.751 &  0.135 & --- & --- & --- & --- \\
\midrule
Teacher & \faCheck & \textcolor{lightgray}{\faTimes} & \textcolor{lightgray}{\faTimes} & --- & --- & \textemdash{} & --- &  --- &  --- & 0.746 & 0.141 & --- & --- \\
\midrule
Ours & \gcell{\faCheck} & \gcell{\faCheck} & \gcell{\faCheck} & \gcell{Discrete} & \gcell{Late} & \gcell{\textcolor{lightgray}{\faTimes}} & \gcell{\textcolor{lightgray}{\faTimes}} & \gcell{0.765} & \gcell{0.138} & \gcell{0.626} & \gcell{0.164} & \gcell{0.766} & \gcell{0.135}\\[0.02cm]
 & \faCheck & \faCheck & \faCheck & Discrete & Late & \faCheck & \textcolor{lightgray}{\faTimes} &  0.740 &  0.149 &  0.628 &  0.165 & 0.745 & 0.145\\
 & \faCheck & \faCheck & \faCheck & Discrete & Late & \textcolor{lightgray}{\faTimes} & \faCheck &  \textbf{0.774} &  0.135 &  0.613 &  0.167 & \textbf{0.778} &  0.130\\
 & \faCheck & \textcolor{lightgray}{\faTimes} & \textcolor{lightgray}{\faTimes} & Discrete & --- & \textcolor{lightgray}{\faTimes} & \textcolor{lightgray}{\faTimes} &  0.673 &  0.152 &  0.637 &  \textbf{0.159} & 0.697 &  0.144\\
 & \faCheck & \faCheck & \faCheck & Discrete & Early & \textcolor{lightgray}{\faTimes} & \textcolor{lightgray}{\faTimes} &  0.741 &  0.141 &  \textbf{0.648} &  0.166 & 0.744 &  0.139\\
\cmidrule(){2-14}
 & \gcell{\faCheck} & \gcell{\faCheck} & \gcell{\faCheck} & \gcell{CoxPH} & \gcell{Late} & \gcell{\textcolor{lightgray}{\faTimes}} & \gcell{\textcolor{lightgray}{\faTimes}} &  \gcell{0.773} &  \gcell{\textbf{0.126}} &  \gcell{0.575} &  \gcell{0.165} & \gcell{0.775} &  \gcell{\textbf{0.126}}\\
 & \faCheck & \faCheck & \faCheck & CoxPH & Late & \faCheck & \textcolor{lightgray}{\faTimes} &  0.743 &  0.144 &  0.625 &  0.165 & 0.746 &  0.141\\
 & \faCheck & \faCheck & \faCheck & CoxPH & Late & \textcolor{lightgray}{\faTimes} & \faCheck &  0.770 &  0.127 &  0.600 & 0.164 & 0.773 & 0.128\\
 & \faCheck & \textcolor{lightgray}{\faTimes} & \textcolor{lightgray}{\faTimes} & CoxPH & --- & \textcolor{lightgray}{\faTimes} & \textcolor{lightgray}{\faTimes} &  0.729 &  0.141 &  0.621 &  0.163 & 0.733 &  0.140\\
& \faCheck & \faCheck & \faCheck & CoxPH & Early & \textcolor{lightgray}{\faTimes} & \textcolor{lightgray}{\faTimes} &  0.753 &  0.135 &  0.611 &  0.168 & 0.755 &  0.134\\
\bottomrule
\vspace{-0.2cm}\\
\multicolumn{9}{l}{\scriptsize $^{*}$For teacher and student models, this comprises $x^{\text{path}}$ and verbalized $x^{\text{cov}}$.}
\end{tabular}
\end{footnotesize}
\label{table:results}
\end{table}
\vspace{-0.1cm}
\paragraph{Qualitative analysis.} We show a positive example of a generated assessment from our model (discrete-time survival model, standard configuration) alongside an excerpt of the corresponding raw pathology report (Fig.~\ref{fig:qualitative}). Consistent evidence spans are highlighted in the same color. Beyond such positive cases, we observe that very long reports can challenge the model: textual coherence may drop and the stated 3-year probability may occasionally be omitted. Additional negative and edge-case examples illustrating these patterns are presented in Appendix~\ref{app:negative_examples}.
\begin{figure}[h]
\centering
\begin{minipage}[t][4cm][t]{0.49\linewidth}
\begin{tcolorbox}[colback=lightgraybg, colframe=lightgrayedge, arc=2mm, boxrule=0.8pt, left=6pt, right=6pt, top=6pt, bottom=6pt]
\parbox[t][3.9cm][t]{\linewidth}{\fontsize{9}{9}\selectfont\raggedright\setlength{\parskip}{0pt}\setlength{\parindent}{0pt}\setlength{\baselineskip}{9pt}%
{\footnotesize\textbf{Raw Pathology Report (excerpt).}}\\[-2pt]
\noindent{\color{lightgrayedge}\rule{\linewidth}{0.5pt}}\\[2pt]
{\ttfamily\fontsize{8}{9}\selectfont
[...] The tumor cells. demonstrate only \greenOne{mild atypia}. \greenTwo{No mitotic figures} are seen [...]. There is \greenThree{no microvascular} \greenThree{proliferation or necrosis}. [...] Scattered \greenFive{MIB-1 reactive cells} are present [...][;] a labeling index of \greenFive{2.5\%} is calculated, consistent with the \greenFive{low grade. histologic} \greenFive{features}. Addendum Diagnosis: \greenOne{Well differentiated} astrocytoma (\greenFour{who grade II}). [...]
}}
\end{tcolorbox}
\end{minipage}\hspace*{0.015\linewidth}
\begin{minipage}[t][4cm][t]{0.49\linewidth}
\begin{tcolorbox}[colback=lightgraybgDarker, colframe=lightgrayedge, arc=2mm, boxrule=0.8pt, left=6pt, right=6pt, top=6pt, bottom=6pt]
\parbox[t][3.9cm][t]{\linewidth}{\fontsize{9}{9}\selectfont\raggedright\setlength{\parskip}{0pt}\setlength{\parindent}{0pt}\setlength{\baselineskip}{9pt}%
{\footnotesize\textbf{Generated Assessment (Student LLM).}}\\[-2pt]
\noindent{\color{lightgrayedge}\rule{\linewidth}{0.5pt}}\\[2pt]
{\ttfamily\fontsize{8}{9}\selectfont
The favorable prognosis is due to the \greenOne{well-differentiated} nature of the tumor (\greenFour{grade II}), the \greenTwo{absence of mitotic} \greenTwo{activity}, and the \greenThree{lack of microvascular} \greenThree{or necrotic features}, indicating a \greenFive{low-grade glioma}. \newline
\\
The estimated 3-year survival probability is: 90\%.
}}
\end{tcolorbox}
\end{minipage}
\caption{Qualitative example: original report (left) and generated assessment (right). Consistent evidence spans are highlighted in the same color.}
\label{fig:qualitative}
\end{figure}
\vspace{-0.3cm}
\section{Conclusion}
\vspace{-0.1cm}
We presented a unified, locally deployable multimodal survival framework that fuses unstructured text (e.g., clinical reports), tabular covariates, and gene expression on a compact LLM backbone with comparatively low hardware requirements. The approach jointly learns calibrated survival estimates and concise, verbalized assessments via a teacher–student scheme. We explored key design choices, including different underlying survival models and fusion strategies. Ongoing work will condition the language model itself on genomics to surface patient-specific, verbalized risk factors (e.g., salient genes or mutations), enabling more trustworthy human–AI collaboration in clinical workflows.

\section*{Acknowledgments}
 MG is supported by the EPSRC Centre for Doctoral Training in Health Data Science (EP/S02428X/1). CY is supported by a UKRI Turing AI Acceleration Fellowship (EP/V023233/1) and received additional funding through EPSRC grant EP/Y018192/1. The results shown here are in whole or part based upon data generated by the TCGA Research Network: \url{https://www.cancer.gov/tcga}.

\bibliography{bibliography}
\bibliographystyle{abbrv}

\newpage
\appendix

\section*{\Large Appendix}
\renewcommand{\thesection}{\Alph{section}}
\renewcommand{\thefigure}{\thesection.\arabic{figure}}
\renewcommand{\thetable}{\thesection.\arabic{table}}
\counterwithin{figure}{section}
\counterwithin{table}{section}

\section{Additional Qualitative Analysis of Generated Assessments}
\label{app:negative_examples}

\begin{figure}[h]
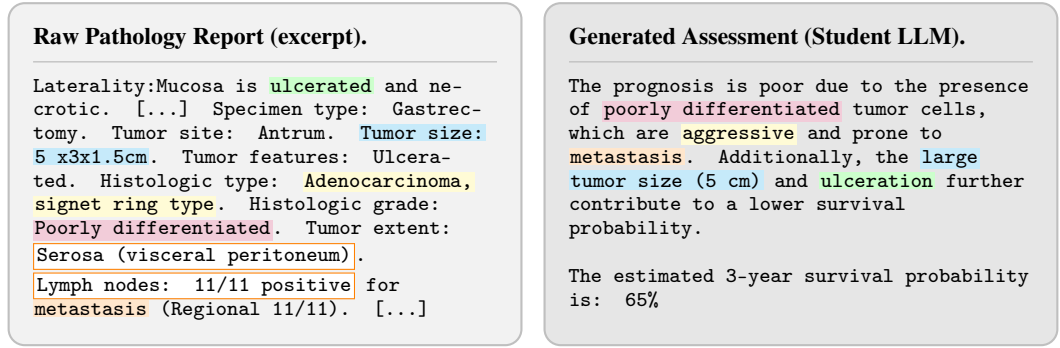

\centering
\begin{minipage}[t][4cm][t]{0.49\linewidth}
\begin{tcolorbox}[colback=lightgraybg, colframe=lightgrayedge, arc=2mm, boxrule=0.8pt, left=6pt, right=6pt, top=6pt, bottom=6pt]
\parbox[t][3.9cm][t]{\linewidth}{\fontsize{9}{9}\selectfont\raggedright\setlength{\parskip}{0pt}\setlength{\parindent}{0pt}\setlength{\baselineskip}{9pt}%
{\footnotesize\textbf{Raw Pathology Report (excerpt).}}\\[-2pt]
\noindent{\color{lightgrayedge}\rule{\linewidth}{0.5pt}}\\[2pt]
{\ttfamily\fontsize{8}{9}\selectfont
Laterality:Mucosa is \greenOne{ulcerated} and ne- crotic. [...] Specimen type: Gastrec- tomy. Tumor site: Antrum. \greenTwo{Tumor size:} \greenTwo{5 x3x1.5cm}. Tumor features: Ulcera- ted. Histologic type: \greenThree{Adenocarcinoma,} \greenThree{signet ring type}. Histologic grade: \greenFive{Poorly differentiated}. Tumor extent: \bad{Serosa (visceral peritoneum)}. \bad{Lymph nodes: 11/11 positive} for \greenFour{metastasis} (Regional 11/11). [...]
}}
\end{tcolorbox}
\end{minipage}\hspace*{0.015\linewidth}
\begin{minipage}[t][4cm][t]{0.49\linewidth}
\begin{tcolorbox}[colback=lightgraybgDarker, colframe=lightgrayedge, arc=2mm, boxrule=0.8pt, left=6pt, right=6pt, top=6pt, bottom=6pt]
\parbox[t][3.9cm][t]{\linewidth}{\fontsize{9}{9}\selectfont\raggedright\setlength{\parskip}{0pt}\setlength{\parindent}{0pt}\setlength{\baselineskip}{9pt}%
{\footnotesize\textbf{Generated Assessment (Student LLM).}}\\[-2pt]
\noindent{\color{lightgrayedge}\rule{\linewidth}{0.5pt}}\\[2pt]
{\ttfamily\fontsize{8}{9}\selectfont
The prognosis is poor due to the presence of \greenFive{poorly differentiated} tumor cells, which are \greenThree{aggressive} and prone to \greenFour{metastasis}. Additionally, the \greenTwo{large} \greenTwo{tumor size (5 cm)} and \greenOne{ulceration} further contribute to a lower survival probability.\newline
\\
The estimated 3-year survival probability is: 65\%
}}
\end{tcolorbox}
\end{minipage}
\caption{Additional qualitative example: original report (left) and student-generated assessment (right). Consistent evidence spans are highlighted with the same color; missing but clinically relevant information is framed in orange. The explanation is reasonable and faithful to the report, but the 3-year survival probability (65\%) is clearly over-optimistic.}
\label{fig:app_qualitative}
\end{figure}

\begin{figure}[h]
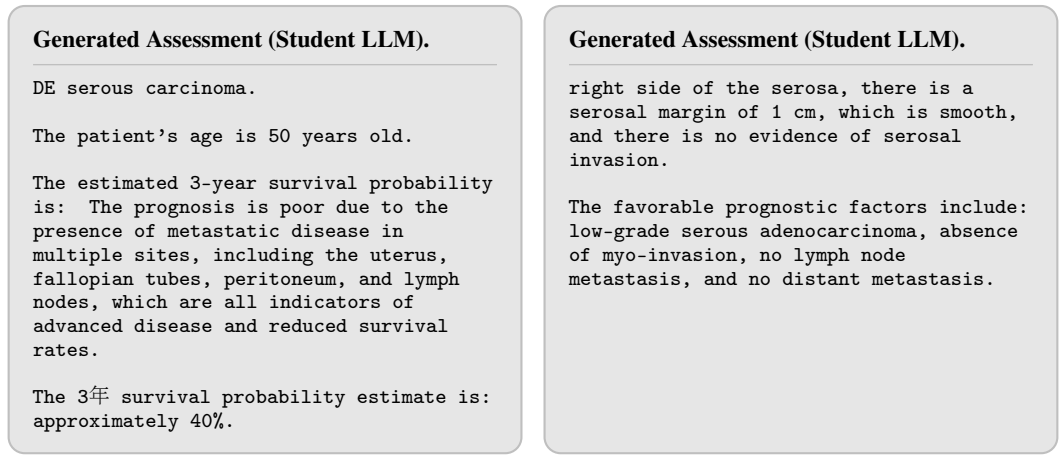

\centering
\begin{minipage}[t][4cm][t]{0.49\linewidth}
\begin{tcolorbox}[colback=lightgraybgDarker, colframe=lightgrayedge, arc=2mm, boxrule=0.8pt, left=6pt, right=6pt, top=6pt, bottom=6pt]
\parbox[t][5.3cm][t]{\linewidth}{\fontsize{9}{9}\selectfont\raggedright\setlength{\parskip}{0pt}\setlength{\parindent}{0pt}\setlength{\baselineskip}{9pt}%
{\footnotesize\textbf{Generated Assessment (Student LLM).}}\\[-2pt]
\noindent{\color{lightgrayedge}\rule{\linewidth}{0.5pt}}\\[2pt]
{\ttfamily\fontsize{8}{9}\selectfont
DE serous carcinoma.\newline

The patient's age is 50 years old. \newline

The estimated 3-year survival probability is: The prognosis is poor due to the presence of metastatic disease in multiple sites, including the uterus, fallopian tubes, peritoneum, and lymph nodes, which are all indicators of advanced disease and reduced survival rates. \newline

The 3\nen{} survival probability estimate is: approximately 40\%.
}}
\end{tcolorbox}
\end{minipage}\hspace*{0.015\linewidth}
\begin{minipage}[t][4cm][t]{0.49\linewidth}
\begin{tcolorbox}[colback=lightgraybgDarker, colframe=lightgrayedge, arc=2mm, boxrule=0.8pt, left=6pt, right=6pt, top=6pt, bottom=6pt]
\parbox[t][5.3cm][t]{\linewidth}{\fontsize{9}{9}\selectfont\raggedright\setlength{\parskip}{0pt}\setlength{\parindent}{0pt}\setlength{\baselineskip}{9pt}%
{\footnotesize\textbf{Generated Assessment (Student LLM).}}\\[-2pt]
\noindent{\color{lightgrayedge}\rule{\linewidth}{0.5pt}}\\[2pt]
{\ttfamily\fontsize{8}{9}\selectfont

right side of the serosa, there is a serosal margin of 1 cm, which is smooth, and there is no evidence of serosal invasion. \newline

The favorable prognostic factors include: low-grade serous adenocarcinoma, absence of myo-invasion, no lymph node metastasis, and no distant metastasis.
}}
\end{tcolorbox}
\end{minipage}
\caption{Two negative examples of student-generated assessments. Both outputs exhibit degraded English fluency; the right example further fails to provide an explicit verbalized probability.}
\label{fig:app_qualitative2}
\end{figure}

While the main paper presents a positive example of a generated assessment (see Fig.~\ref{fig:qualitative}), here we include additional negative examples to illustrate common failure modes. We find that the model struggles particularly with longer pathology reports, where fluency can degrade and the numeric prediction may become less reliable. In Fig.~\ref{fig:app_qualitative}, the explanation remains sensible and grounded in the report but the 3-year probability is over-optimistic. Fig.~\ref{fig:app_qualitative2} shows two negative cases with poor English fluency; the right example additionally fails to produce an explicit verbalized probability.

\section{Data and Preprocessing}
\label{app:data}
The Cancer Genome Atlas (TCGA) \cite{TCGA2013} is a large multi‑cancer resource providing harmonized clinical, pathology, and genomics data. We assembled a cohort pairing survival outcomes with three inputs per sample: an unstructured pathology report, structured covariates, and gene‑expression profiles. Survival times were administratively censored after 5 years. Covariates were processed following a data preparation strategy inspired by \cite{jeanselme2024}.\footnote{\url{https://github.com/Jeanselme/LLM-For-Survival-Analysis}} We included age at initial pathologic diagnosis, sex, race, AJCC pathologic tumor stage (I, II, III), and cancer type. Age and stage were min–max scaled. Sex and race were encoded as binary indicators. Cancer type was grouped into families (gastrointestinal, gynecological, genitourinary, respiratory, skin, brain, and other) and represented via one‑hot indicators. Gene‑expression profiles comprised 20,531 genes; within‑gene missing entries were imputed to 0. Samples with other missing critical information were excluded (survival outcome, clinical report, gene expression, or covariates). The remaining 8,902 samples were split into training/validation/test sets in proportions 70/10/20\%.
 
\section{Implementation Details}
\label{app:impl}
\subsection{Details on Teacher Pipeline and Student Input Sequence Construction}
\label{app:teacher_pipeline}
We construct the student’s training target with a two‑step teacher pipeline (Fig.~\ref{fig:teacher_pipeline}). First, a larger teacher LLM is queried in three prompts with numeric‑only instructions to return survival probabilities at 1, 3, and 5 years (same horizons as in \cite{jeanselme2024}) from the input text consisting of the pathology report with a short patient‑information snippet. We extract the numeric probabilities using regular expressions, interpolate any missing probabilities (if all three probabilities are missing, we impute them using the corresponding means), and then fit a simple parametric survival curve. This multi‑horizon fitting grounds the 3‑year estimate in three independent numeric prompts rather than a single query, improving robustness to prompt variance. We explored Weibull \cite{Weibull1951}, log-logistic, spline, and exponential curves and found that a simple exponential fit performed best in terms of IBS and $C^{\text{td}}$. We therefore used the exponential model throughout to estimate the 3-year survival probability for the second step, in which the teacher is prompted to generate a concise prognosis explanation conditioned on the same text and the 3‑year survival percentage (rounded to the nearest 5\%). The student model learns a target sequence that concatenates the explanation with a marked sentence verbalizing the 3‑year probability, delimited by \texttt{<<VPROB>>} and \texttt{<<END\_VPROB>>}, enabling span‑ and number‑weighted language‑model losses during training.

\subsection{Late Fusion Blending}
\label{app:latefusion}
We present the precise late fusion blending used for discrete-time and CoxPH heads. For discrete heads, the learnable modality gates are per–time-bin vectors $\gamma^{\text{text}},\gamma^{\text{cov}},\gamma^{\text{ge}}\in[0,1]^B$; for CoxPH, the modality gates are scalars $\gamma^{\text{text}},\gamma^{\text{cov}},\gamma^{\text{ge}}\in[0,1]$. The discrete logits $o$ and CoxPH scores $g$ are then given by
\begin{equation}
\begin{aligned}
 o \; &=\; (1-\gamma^{\text{ge}})\,\big[(1-\gamma^{\text{cov}})\,o^{\text{text}}+\gamma^{\text{cov}}\,o^{\text{cov}}\big] \;+\; \gamma^{\text{ge}}\,o^{\text{ge}},\\
 g \; &=\; (1-\gamma^{\text{ge}})\,\big[(1-\gamma^{\text{cov}})\,g^{\text{text}}+\gamma^{\text{cov}}\,g^{\text{cov}}\big] \;+\; \gamma^{\text{ge}}\,g^{\text{ge}}.
\end{aligned}
\end{equation}

\subsection{Metrics Computation}
\label{app:metrics}
We report time-dependent concordance \(C^{\text{td}}\) and integrated Brier score (IBS) calculated as 
\begin{gather}
C^{\text{td}} = \mathbb{P}\!\left(\hat{S}(t_i\mid x_i) < \hat{S}(t_i\mid x_j)\;\middle|\; t_i < t_j,\; e_i=1\right)\\
\text{IBS} = \frac{1}{t_{\max}} \int_{0}^{t_{\max}} \frac{1}{N} \sum_{i=1}^{N}
\left[ \frac{\hat{S}(t\mid x_i)^2\, \mathds{1}\{t_i \le t,\; e_i=1\}}{\hat{G}(t_i)}
+ \frac{\big(1-\hat{S}(t\mid x_i)\big)^2\, \mathds{1}\{t_i > t\}}{\hat{G}(t)} \right] dt,
\end{gather}
$\text{with } t_{\max} = \max_i t_i$. $\hat{G}(\cdot)$ is the Kaplan--Meier estimate of the censoring survival function used for inverse-probability-of-censoring weighting (IPCW) \cite{Kvamme2019_pycox}. 

\subsection{Model Configurations, Hyperparameters, and Computational Details}
The configurations and hyperparameter settings of all models are provided in Table~\ref{tab:hyperparams}. 
Specifically, we implement both survival heads and the autoencoder as MLPs. We use distinct learning rates across parameters in different model components (LLM, survival head, autoencoder, and gating parameters). In particular, the LLM’s learning rate is set lower to avoid overwriting pre-trained knowledge. To reduce memory consumption, we freeze all but the last 18 layers of the student LLM during fine-tuning, truncate pathology reports to 820 tokens, and use a small batch size.

For baselines, because no public BERTSurv implementation is available, we re-implemented it in‑house following \cite{Zhao2021_BERTSurv} and applied the reported hyperparameters. As BERTSurv does not directly handle gene‑expression inputs, we trained the same gene‑expression autoencoder offline to obtain fixed latents, which we then concatenated to the ClinicalBERT text embedding (together with other tabular covariates). For the covariates‑only and gene‑expression‑only baselines, we reused the same survival‑model configuration as in our framework to ensure comparability. In CoxPH settings, we compute the Breslow baseline hazards \cite{Breslow1975} on the combined training and validation sets and use them for test‑time evaluation.

\begin{table}[t!]
\centering
\setlength{\tabcolsep}{7.5pt}
\caption{Hyperparameters of student model, teacher model, and baselines.}
\label{tab:hyperparams}
\begin{small}
\begin{NiceTabular}{>{\centering\arraybackslash}p{1cm} l l l}
\toprule
\textbf{Model} & \textbf{Component} & \textbf{Parameter} & \textbf{Value}\\
\midrule
\Block[draw=gray, fill=gray!15]{28-1}{\rotatebox{90}{\textbf{Student Model}}} & LLM & LLM Name & DeepSeek-R1 Distill Qwen-1.5B \\
&  & Freezing & All but last 18 transformer layers \\
&  & Precision & bfloat16 \\
&  & Attention dropout & 0.1 \\
&  & Learning rate & 5e-5\\
\cmidrule(l){2-4}
& Autoencoder & Latent dim & 128 \\
&  & Encoder layers & [4096, 2048, 1024, 512, 256] \\
&  & Activation & ReLU\\
&  & Dropout & 0.3  \\
&  & $\alpha$ (CoxPH / Discrete) & 1e-8 / 1e-9 \\
&  & Learning rate & 1e-3 \\
\cmidrule(l){2-4}
& Survival Model & Layers & [100,100,100]\\
&  & Activation & ReLU\\
&  & Dropout & 0.3 \\
&  & Time bins $B$ (Discrete) & 30 \\
&  & $\beta$ (CoxPH / Discrete) & 5.0 / 1.0\\ 
&  & Learning rate & 1e-3 (gates 1e-4)\\ 
\cmidrule(l){2-4}
& Optimization & Batch size & 16 (512 under pre-training)\\
&  & Epochs / Patience & 30 / 5 (1000 / 5 under pre-training)\\
&  & Optimizer & AdamW (weight decay 0.01) \\
&  & Span weights & sentence $w$=2.0; number $w^{\text{num}}$=5.0 \\
\cmidrule(l){2-4}
& Generation Parameters & Temperature & 0.3\\
&  & Min / max \# of new tokens & 50 / 350 \\
&  & \# Beams & 3\\
&  & Top-k & 20\\
&  & Top-p & 0.9\\
&  & Repetition penalty & 1.3\\
&  & No-repeat-ngram-size & 3\\ 
\midrule
\Block[draw=gray, fill=gray!15]{4-1}{\rotatebox{90}{\parbox{1cm}{\centering \textbf{Teacher}\\\textbf{Model}}}} & Generation Parameters & LLM Name & DeepSeek-R1 Distill Qwen-32B \\
&  & Temperature & 0.1 \\
&  & Max \# of new tokens & 10 / 300 \\
&  & (Num. / Expl. prompt) &  \\
\midrule
\Block[draw=gray, fill=gray!15]{9-1}{\rotatebox{90}{\textbf{BERTSurv}}} & LLM & LLM Name& ClinicalBERT \\
\cmidrule(l){2-4}
& Survival Model & Type & CoxPH \\
&  & Layers & [100,100,100] \\
&  & Activation & SELU \\
&  & Dropout & 0.1 \\
\cmidrule(l){2-4}
& Optimization & Batch size & 24\\
& & Epochs / Patience & 30 / 5\\
&  & LR & 1e-2 \\
&  & Optimizer & Adam (weight decay 0.01) \\
\midrule
\Block[draw=gray, fill=gray!15]{5-1}{\rotatebox{90}{\parbox{2cm}{\centering \textbf{Cov-only /}\\\textbf{GE-only}}}} & Survival Model & \multicolumn{2}{l}{AS IN STUDENT MODEL} \\
\cmidrule(l){2-4}
& Optimization & Batch size & 512\\
& & Epochs / Patience & 1000 / 5\\
&  & LR & 1e-3 \\
&  & Optimizer & AdamW (weight decay 0.01) \\
\bottomrule
\end{NiceTabular}
\end{small}
\vspace{-0.5cm}
\end{table}
All experiments were performed on a computer cluster with one NVIDIA A100 (80 GB) GPU, six CPU cores, and 20 GB system RAM per task. At inference, the student model’s resource footprint is substantially lower; evaluation can run on smaller GPUs or even CPU‑only, supporting local deployability.

\end{document}